%% file: acl2021.tex
\pdfoutput=1
\documentclass[11pt,a4paper]{article}
\usepackage[hyperref]{acl2021}
\usepackage{times}
\usepackage{latexsym}
\usepackage{graphicx}
\usepackage{bm}
\usepackage{multirow} 
\usepackage{amsmath}
\usepackage{amssymb}
\usepackage{amsfonts}
\usepackage{booktabs}
\usepackage{adjustbox}
\usepackage{xspace}
\usepackage{cleveref}
\usepackage{dashbox}
\usepackage{IEEEtrantools}

\usepackage{microtype}

\aclfinalcopy 
\def\aclpaperid{403} 

\newcommand{\modelname}{\textsc{Git}\xspace}
\newcommand\dboxed[1]{\dbox{\strut{#1}}}

\title{Document-level Event Extraction via Heterogeneous \\ Graph-based Interaction Model with a Tracker}

\author{
  Runxin Xu$^{1}$, 
  Tianyu Liu$^{1}$, 
  Lei Li$^{3}$ \and
  Baobao Chang$^{1,2}$\footnotemark[1]
  \\
  $^{1}$Key Laboratory of Computational Linguistics, Peking University, MOE, China\\
  $^{2}$Peng Cheng Laboratory, Shenzhen, China\\
  $^{3}$ByteDance AI Lab\\
  \texttt{
     runxinxu@gmail.com,lileilab@bytedance.com
  } \\
  \texttt{
     \{tianyu0421,chbb\}@pku.edu.cn
  }
}

\date{}

\begin{document}
\maketitle

\renewcommand{\thefootnote}{\fnsymbol{footnote}} 
\footnotetext[1]{Corresponding author.}

\input{main/abstract}
\input{main/introduction}

\input{main/formulation}

\input{main/model}
\input{main/experiments}
\input{main/related}
\input{main/conclusion}

\section*{Acknowledgments}
The authors would like to thank Changzhi Sun, Mingxuan Wang, and the anonymous reviewers for their thoughtful and constructive comments.
This paper is supported in part by the National Key R\&D Program of China under Grand No.2018AAA0102003, the National Science Foundation of China under Grant No.61936012 and 61876004.

\bibliographystyle{acl_natbib}
\bibliography{acl2021}

\appendix
\input{main/appendix.tex}

\end{document}

%% file: main/abstract.tex
\begin{abstract}

Document-level event extraction aims to recognize event information from a whole piece of article.
Existing methods are not effective due to two challenges of this task: a) the target event arguments are scattered across sentences; b) the correlation among events in a document is non-trivial to model.
In this paper, we propose Heterogeneous \textbf{G}raph-based \textbf{I}nteraction Model with a \textbf{T}racker (\modelname) to solve the aforementioned two challenges.
For the first challenge, \modelname constructs a heterogeneous graph interaction network to capture global interactions among different sentences and entity mentions.
For the second, \modelname introduces a \textit{Tracker} module to track the extracted events and hence capture the interdependency among the events.
Experiments on a large-scale dataset  \citep{zheng-etal-2019-doc2edag} show \modelname outperforms the previous methods by 2.8 F1.
Further analysis reveals \modelname is effective in extracting multiple correlated events and event arguments that scatter across the document.
Our code is available at \url{https://github.com/RunxinXu/GIT}.

\end{abstract}

%% file: main/introduction.tex
\section{Introduction}
\begin{figure}
    \centering
    \includegraphics[width=0.95\linewidth]{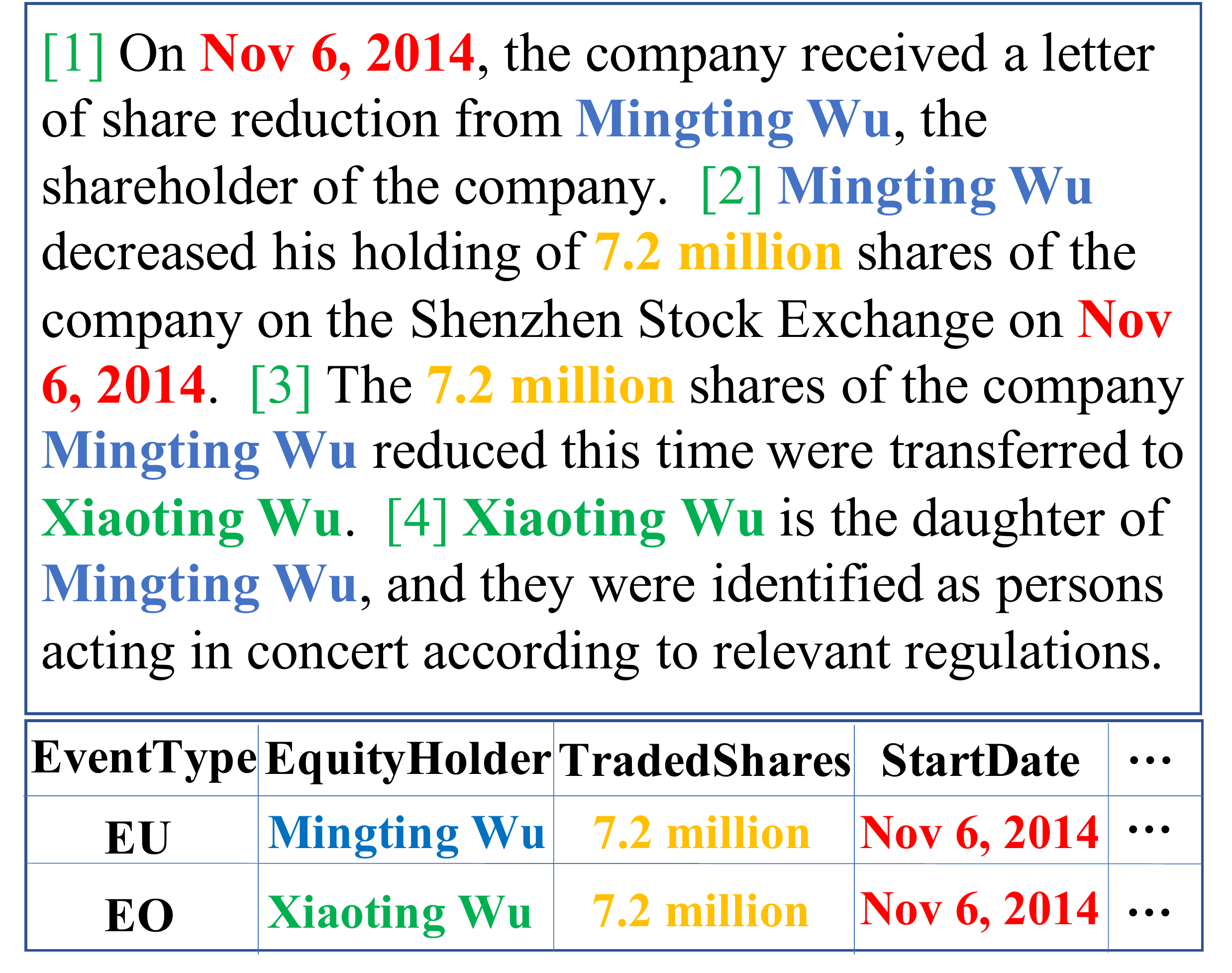}
    \caption{An example document from a Chinese dataset proposed by \citet{zheng-etal-2019-doc2edag} in the financial domain, and we translate it into English for illustration. Entity mentions are colored. Due to space limitation, we only show four associated sentences and three argument roles of each event type. The complete original document can be found in Appendix~\ref{appendix-running-example}. EU: Equity Underweight, EO: Equity Overweight.}
    \label{fig:running-example}
\end{figure}

Event Extraction (EE) is one of the key and challenging tasks in Information Extraction (IE), which aims to detect events and extract their arguments from the text.
Most of the previous methods \citep{ chen-etal-2015-event, nguyen-etal-2016-joint, liu-etal-2018-jointly, yang-etal-2019-exploring-pre, du-cardie-2020-event} focus on sentence-level EE, extracting events from a single sentence.
The sentence-level model, however, fails to extract events whose arguments locate in different sentences, which is much more common in real-world scenarios.
Hence, extracting events at the document-level is critical and has attracted much attention recently \citep{yang-etal-2018-dcfee,zheng-etal-2019-doc2edag,du-cardie-2020-document, DBLP:journals/corr/abs-2008-09249}.

Though promising, document-level EE still faces two critical challenges.
\textbf{Firstly}, the arguments of an event record may scatter across sentences, which requires a comprehensive understanding of the cross-sentence context.
Figure~\ref{fig:running-example} illustrates an example that one \textit{Equity Underweight} (EU) and one \textit{Equity Overweight} (EO) event records are extracted from a financial document.
It is less challenging to extract the EU event because all the related arguments appear in the same sentence (Sentence $2$).
However, for the arguments of EO record, \textit{Nov 6, 2014} appears in Sentence $1$ and $2$ while \textit{Xiaoting Wu} in Sentence $3$ and $4$.
It would be quite challenging to identify such events without considering global interactions among sentences and entity mentions.
\textbf{Secondly}, a document may express several correlated events simultaneously, and recognizing the interdependency among them is fundamental to successful extraction.
As shown in Figure~\ref{fig:running-example}, the two events are interdependent because they correspond to exactly the same transaction and therefore share the same \textit{StartDate}.
Effective modeling on such interdependency among the correlated events remains a key challenge in this task.


\citet{yang-etal-2018-dcfee} extracts events from a central sentence and query the neighboring sentences for missing arguments, which ignores the cross-sentence correspondence between augments.
Though \citet{zheng-etal-2019-doc2edag} takes a first step to fuse the sentences and entities information via Transformer, they neglect the interdependency among events. 
Focusing on single event extraction, \citet{du-cardie-2020-document} and \citet{DBLP:journals/corr/abs-2008-09249} concatenate multiple sentences and only consider a single event, which lacks the ability to model multiple events scattered in a long document.

To tackle the aforementioned two challenges, in this paper, we propose a Heterogeneous \textbf{G}raph-based \textbf{I}nteraction Model with a \textbf{T}racker (\modelname) for document-level EE.
To deal with scattered arguments across sentences, we focus on the \emph{Global Interactions} among sentences and  entity mentions.
Specifically, we construct a heterogeneous graph interaction network with mention nodes and sentence nodes, and model the interactions among them by four types of edges (i.e., sentence-sentence edge, sentence-mention edge, intra-mention-mention edge, and inter-mention-mention edge) in the graph neural network.
In this way, \modelname jointly models the entities and sentences in the document from a global perspective.

To facilitate the multi-event extraction, we target on the \emph{Global Interdependency} among correlated events.
Concretely we propose a \textit{Tracker} module to continually tracks the extracted event records with a global memory. 
In this way, the model is encouraged to incorporate the interdependency with other correlated event records while predicting.

We summarize our contributions as follows:
\begin{itemize}
    \item We construct a heterogeneous graph interaction network for document-level EE. 
    With different heterogeneous edges,
    the model could capture the global context for the scattered event arguments across different sentences.
    \item We introduce a novel \textit{Tracker} module to track the extracted event records. The \textit{Tracker} eases the difficulty of extracting correlated events, as interdependency among events would be taken into consideration.
    \item Experiments show \modelname outperforms the previous state-of-the-art model by $2.8$ F1 on the large-scale public dataset \citep{zheng-etal-2019-doc2edag} with $32,040$ documents, especially on cross-sentence events and multiple events scenarios (with $3.7$ and $4.9$ absolute increase on F1).
\end{itemize}

%% file: main/formulation.tex
\section{Preliminaries}
We first clarify some important notions. 
a) \textbf{entity mention}: a text span within document that refers to an entity object; 
b) \textbf{event argument}: an entity playing a specific event role. Event roles are predefined for each event type;
c) \textbf{event record}: an entry of a specific event type containing arguments for different roles in the event. For simplicity, we use \textbf{record} for short in the following sections.

Following \citet{zheng-etal-2019-doc2edag}, given a document composed of sentences $\mathcal{D}=\left\{s_i\right\}^{\left| \mathcal{D} \right |}_{i=1}$ and a sentence containing a sequence of words $s_i=\left\{w_j\right\}^{\left | s_i \right |}_{j=1}$, the task aims to handle three sub-tasks : 
1) \textbf{entity extraction}: extracting entities $\mathcal{E}=\left\{e_i\right\}^{\left | \mathcal{E} \right |}_{i=1}$ from the document to serve as argument candidates. 
An entity may have multiple mentions across the document.
2) \textbf{event types detection}: detecting specific event types that are expressed by the document.
3) \textbf{event records extraction}: finding appropriate arguments for the expressed events from entities, which is the most challenging and also the focus of our paper.
The task does not require to identify event triggers~\citep{DBLP:conf/aaai/ZengFMWYSZ18, liu-etal-2019-event}, which reduces manual effort of annotation and the application scenarios becomes more extensive.

%% file: main/model.tex
\section{Methodology}

\begin{figure*}
    \centering
    \includegraphics[width=1.0\linewidth]{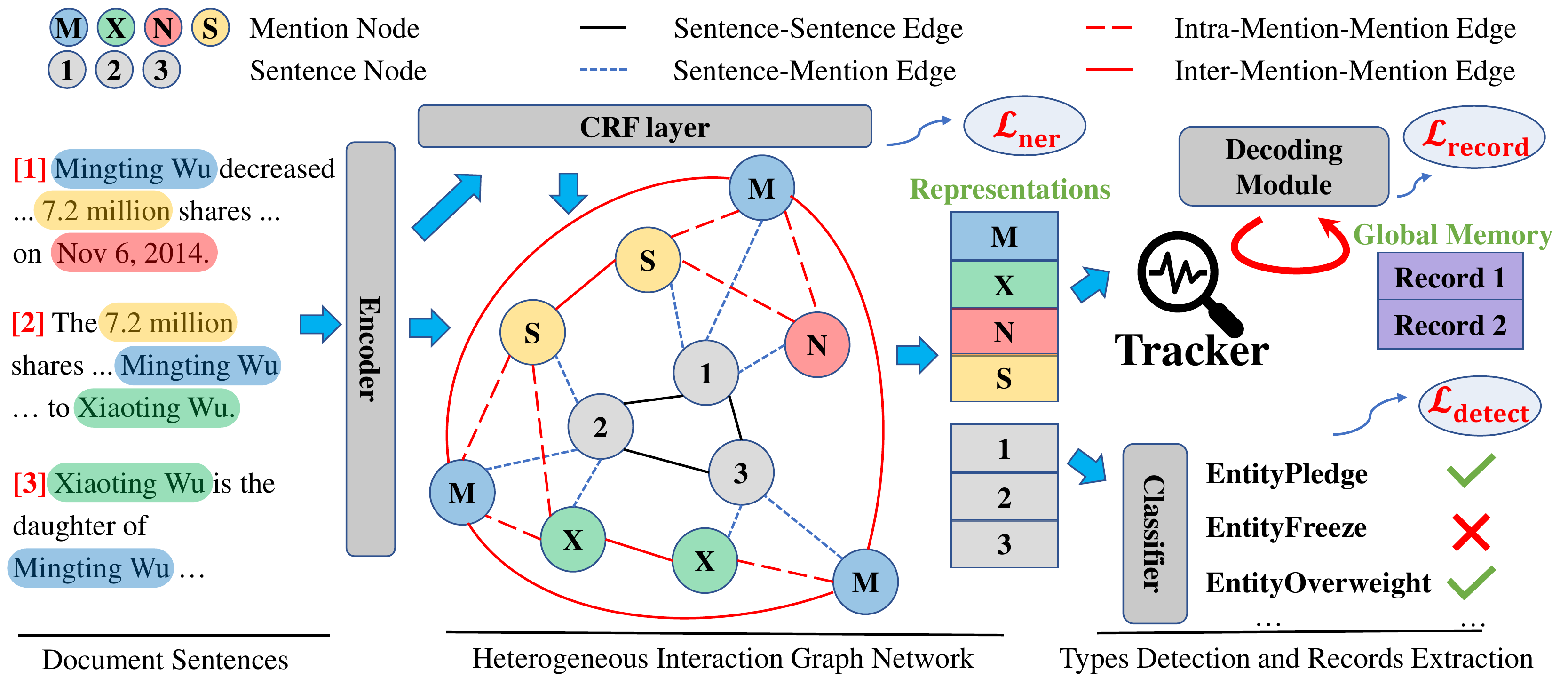}
    \caption{Overview of our \modelname. Firstly, sentences of the document are fed into the encoder to obtain contextualized representation, followed by a CRF layer to extract entities. Then \modelname constructs a heterogeneous graph interaction network with mention nodes and sentence nodes, which captures the global interactions among them based on GCNs.
    After obtaining document-aware representations of entities and sentences, \modelname detects event types and extracts records through the decoding module with a \textit{Tracker}. 
    The \textit{Tracker} tracks extracted records with global memory, based on which the decoding module incorporates global interdependency among correlated event records. Different entities are marked by different colors. M: Mingting Wu. X: Xiaoting Wu. N: Nov 6, 2014. S: 7.2 million.}
    \label{fig:model}
\end{figure*}

As shows in Figure~\ref{fig:model}, \modelname first extracts candidate entities through sentence-level neural extractor (Sec \ref{sec:entity_extraction}). 
Then we construct a heterogeneous graph to model the interactions among sentences and entity mentions  (Sec  \ref{sec:git_structure}), and detect event types expressed by the document (Sec \ref{sec:detect_events}).
Finally we introduce a \textit{Tracker} module to continuously track all the records with global memory, in which we utilize the global interdependency among records for multi-event extraction (Sec \ref{sec:records_extraction}).

\subsection{Entity Extraction}
\label{sec:entity_extraction}

Given a sentence $s = \{ w_j \}_{j=1}^{|s|}\in \mathcal{D}$, we encode $s$ into a sequence of vectors
$\left\{g_{j}\right\}^{\left | s_i \right |}_{j=1}$ using Transformer~\citep{NIPS2017_7181}:
\begin{equation*}
\{ g_{1}, \ldots, g_{|s|}\} = \texttt{Transformer}(\{ w_1, \ldots, w_{|s|} \})
\end{equation*}
The word representation of $w_j$ is a sum of the corresponding token and position embeddings.

We extract entities at the sentence level and formulate it as a sequence tagging task with BIO (Begin, Inside, Other) schema.
We leverage a conditional random field (CRF) layer to identify entities.
For training, we minimize the following loss:
\begin{equation}
\setlength\abovedisplayskip{8pt}
\setlength\belowdisplayskip{8pt}
\mathcal{L}_{\mathrm{ner}} = -\sum_{s \in \mathcal{D}} \log P(y_s|s)
\label{loss:ner}
\end{equation}
where $y_s$ is the golden label sequence of $s$.
For inference, we use Viterbi algorithm to decode the label sequence with the maximum probability.

\subsection{Heterogeneous Graph Interaction Network}
\label{sec:git_structure}

An event may span multiple sentences in the document, which means its corresponding entity mentions may also scatter across different sentences. 
Identifying and modeling these entity mentions in the cross-sentence context is fundamental in document EE. 
Thus we build a heterogeneous graph $\mathcal{G}$ which contains entity mention nodes and sentence nodes in the document $\mathcal{D}$.
In the graph $\mathcal{G}$, 
interactions among multiple entity mentions and sentences can be explicitly modeled.
For each entity mention node $e$, 
we initialize node embedding $h_e^{(0)} = \texttt{Mean}(\{g_j\}_{j \in e}) $ by averaging the representation of the contained words.
For each sentence node $s$,
we initialize node embedding $h_s^{(0)} = \texttt{Max}(\{g_j\}_{j \in s}) + \texttt{SentPos}(s)$ by max-pooling all the representation of words within the sentence plus sentence position embedding.

\emph{To capture the interactions among sentences and mentions}, we introduce four types of edges.
    \paragraph{Sentence-Sentence Edge (S-S)} Sentence nodes are fully connected to each other with S-S edges. 
    In this way, we can easily capture the global properties in the document with sentence-level interactions, e.g., the long range dependency between any two separate sentences in the document would be modeled efficiently with S-S edges.
    \paragraph{Sentence-Mention Edge (S-M)} We model the local context of an entity mention in a specific sentence with S-M edge, specifically the edge connecting the mention node and the sentence node it belongs to.
    \paragraph{Intra-Mention-Mention Edge (M-M$_{\mathrm{intra}}$)} We connect distinct entity mentions in the same sentences with M-M$_{\mathrm{intra}}$ edges. 
    The co-occurrence of mentions in a sentence indicates those mentions are likely to be involved in the same event. 
    We explicitly model this indication by M-M$_{\mathrm{intra}}$ edges.
    \paragraph{Inter-Mention-Mention Edge (M-M$_{\mathrm{inter}}$)}The entity mentions that corresponds to the same entity are fully connected with each other by M-M$_{\mathrm{inter}}$ edges. 
    As in document EE, an entity usually corresponds to multiple mentions across sentences, we thus use M-M$_{\mathrm{inter}}$ edge to track all the appearances of a specific entity, which facilitates the long distance event extraction from a global perspective. 

In Section.~\ref{sec:analysis}, experiments show that all of these four kinds of edges play an important role in event detection, and the performance would decrease without any of them.

After heterogeneous graph construction 
\footnote{Traditional methods in sentence-level EE also utilize graph to extract events \citep{liu-etal-2018-jointly, yan-etal-2019-event}, based on the dependency tree.
However, our interaction graph is heterogeneous and have no demands for dependency tree.
}, 
we apply multi-layer Graph Convolution Network \citep{kipf2017semi} to model the global interactions. 
Given node $u$ at the $l$-th layer, the graph convolutional operation is defined as follows:
\begin{equation*}
       h_{u}^{(l + 1)} = \texttt{ReLU} \left(\sum_{k\in\mathcal{K}}\sum_{v\in\mathcal{N}_k(u) \bigcup \{u\}} \frac{1}{c_{u,k}} W^{(l)}_k h_{v}^{(l)} \right)
\end{equation*}
where $\mathcal{K}$ represents different types of edges, $W^{(l)}_k\in \mathbb{R}^{d_m \times d_m}$ is trainable parameters.
$\mathcal{N}_k(u)$ denotes the neighbors for node $u$ connected in $k$-th type edge and $c_{u,k}$ is a normalization constant.
We then derive the final hidden state $h_u$ for node $u$,
\begin{equation*}
    h_u = W_a[h_{u}^{(0)}; h_{u}^{(1)}; \ldots; h_{u}^{(L)}]
\end{equation*}
where $h^{(0)}_u$ is the initial node embedding of node $u$, and $L$ is the number of GCN layers.

Finally, we obtain the sentence embedding matrix $S = [h_1^\top \ h_2^\top \ \dots \  h_{\left| \mathcal{D} \right |}^\top] \in \mathbb{R}^{d_m \times \left| \mathcal{D} \right |}$ and entity embedding matrix $E \in \mathbb{R}^{d_m \times {\left | \mathcal{E} \right |}}$. 
The $i$-th entity may have many mentions, where we simply use string matching to detect entity coreference following ~\citet{zheng-etal-2019-doc2edag}
, and the entity embedding $E_i$ is computed by the average of its mention node embedding, $E_i = \texttt{Mean}(\{h_j \}_{j \in \mathrm{Mention}(i)})$.
In this way, the sentences and entities are interactively represented in a context-aware way.

\subsection{Event Types Detection}
\label{sec:detect_events}
Since a document can express events of different types, we formulate the task as a multi-label classification and leverage sentences feature matrix $S$ to detect event types:
\begin{equation*}
\begin{split}
    A &= \texttt{MultiHead}(Q, S, S) \in \mathbb{R}^{d_m \times T} \\
    R &= \texttt{Sigmoid}(A^\top W_t) \in \mathbb{R}^{T}
\end{split}
\end{equation*}
where $Q \in \mathbb{R}^{d_m \times T}$ and $W_t \in \mathbb{R}^{d_m}$ are trainable parameters, and $T$ denotes the number of possible event types.
$\texttt{MultiHead}$ refers to the standard multi-head attention mechanism with Query/Key/Value.
Therefore, we derive the event types detection loss with golden label $\widehat{R} \in \mathbb{R}^{T} $:
\begin{equation}
\setlength\abovedisplayskip{8pt}
\setlength\belowdisplayskip{8pt}
\begin{split}
    \mathcal{L}_{\mathrm{detect}} =& - \sum_{t=1}^{T} \mathbb{I} \left ( \widehat{R}_t=1 \right ) \log P \left ( R_t|\mathcal{D} \right ) \\ 
     & + \mathbb{I} \left ( \widehat{R}_t=0 \right ) \log \left(  1 - P \left ( R_t |\mathcal{D}\right ) \right )
\end{split}
\label{loss:detect}
\end{equation}

\subsection{Event Records Extraction}
\label{sec:records_extraction}
\begin{figure}[t]
    \centering
    \includegraphics[width=1.0\linewidth]{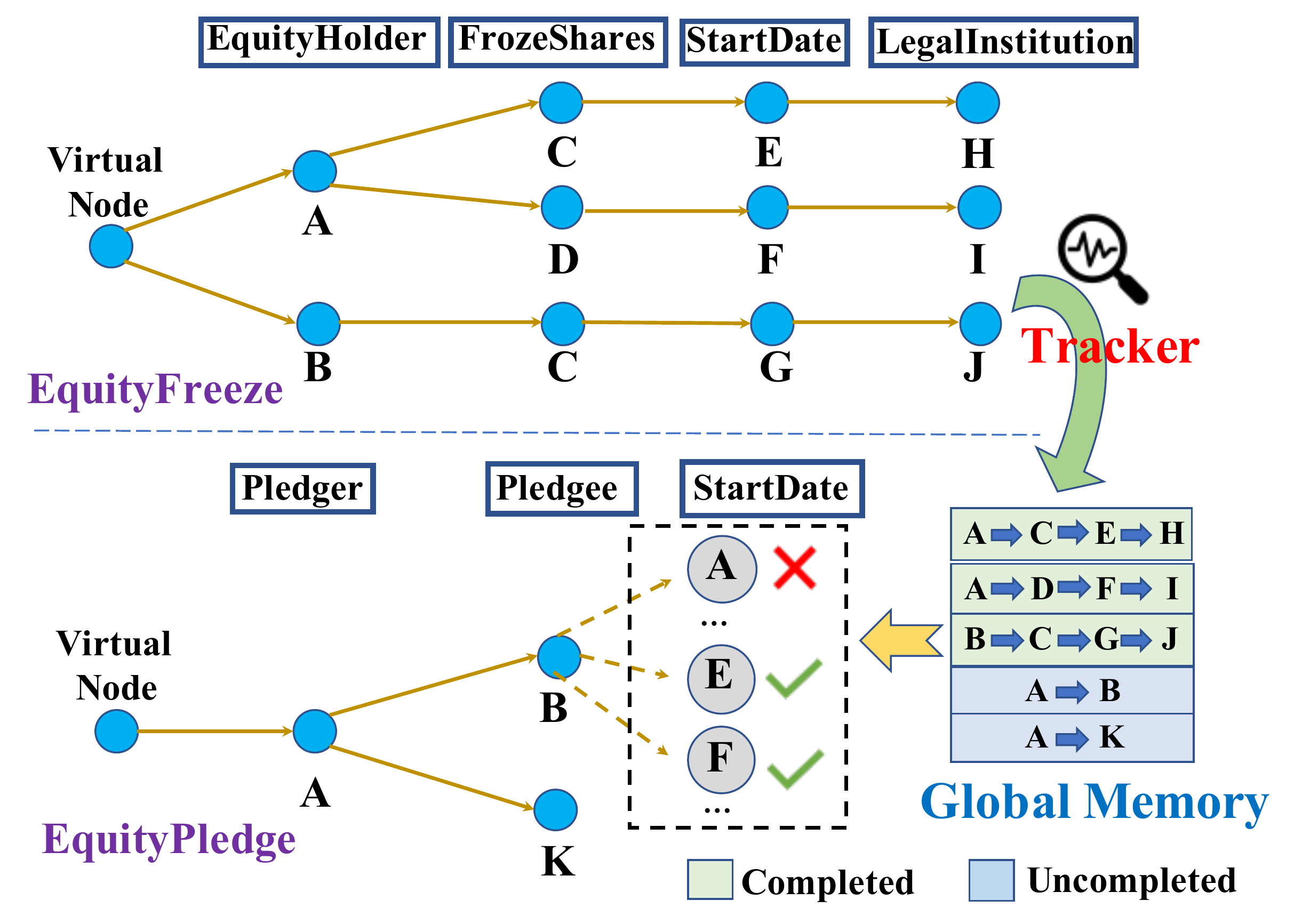}
    \caption{The decoding module of \modelname.
    Three \textit{Equity Freeze} records have been extracted completely, and \modelname is predicting the \textit{StartDate} role for the \textit{Equity Pledge} records (in the dashed frame \dboxed{} ), based on the global memory where \textit{Tracker} tracks the records on-the-fly. 
    Both entity E and F are predicted as the legal \textit{StartDate} role while A is not.
    Pre-defined argument roles are shown in the blue box, and \modelname extracts records in this order.
    Capital letters (A-K) refer to different entities. 
    A path from root to leaf node represents one unique event record.
    }
    \label{fig:tracker}
\end{figure}

Since a document is likely to express multiple event records and the number of records cannot be known in advance, we decode records by  expanding a tree orderly as previous methods did \citep{zheng-etal-2019-doc2edag}. 
However, they treat each record independently.
Instead, \emph{to incorporate the interdependency among event records}, we propose a \textit{Tracker} module, which improves the model performance.

To be self-contained, we introduce the ordered tree expanding in this paragraph.
In each step, we extract event records of a specific event type. 
The arguments extraction order is predefined so that the extraction is modeled as a constrained tree expanding task\footnote{We simply adopt the order used by \citet{zheng-etal-2019-doc2edag}.}.
Taking \textit{Equity Freeze} records as an example, as shown in Figure~\ref{fig:tracker}, we firstly extract \textit{EquityHolder}, followed by \textit{FrozeShares} and others.
Starting from a virtual root node, the tree expands by predicting arguments in a sequential order. 
As there may exist multiple eligible entities for the event argument role, the current node will expand several branches during extraction, with different entities assigned to the current role.
This branching operation is formulated as \emph{multi-label} classification task.
In this way, each path from the root node to the leaf node is identified as a unique event record.

Interdependency exists extensively among different event records.
For example, as shown in Figure~\ref{fig:running-example}, an \textit{Equity Underweight} event record is closely related to an \textit{Equity Overweight} event record, and they may share some key arguments or provide useful reasoning information.
To take advantage of such interdependency, we propose a novel \textit{Tracker} module inspired by memory network \citep{DBLP:journals/corr/WestonCB14}.
Intuitively, the \textit{Tracker} continually tracks the extracted records on-the-fly and store the information into a global memory.
When predicting arguments for current record, the model will query the global memory and therefore make use of useful interdependency information of other records.

In detail, for the $i$-th record path consisting of a sequence of entities, the \textit{Tracker} encodes the corresponding entity representation sequence $U_i = [E_{i1}, E_{i2}, ...]$ into an vector $G_i$ with an LSTM (last hidden state) and add event type embedding.
Then the compressed record information is stored in the global memory $G$, which is shared across different event types as shown in Figure~\ref{fig:tracker}.
For extraction, given a record path $U_i \in \mathbb{R}^{d_m \times (J-1)}$ with the first $J-1$ arguments roles, we predict the $J$-th role by injecting role-specific information into entity representations, $\overline{E} = E + \texttt{Role}_J$, where $\texttt{Role}_J$ is the role embedding for the $J$-th role.
Then we concatenate $\overline{E}$, sentences feature $S$, current entities path $U_i$, and the global memory $G$, followed by a transformer to obtain new entity feature matrix $\widetilde{E} \in \mathbb{R}^{d_m \times \left | \mathcal{E} \right |}$, which contains global role-specific information for all entity candidates.\footnote{To distinguish different parts in the concatenated vector, we also add segment embedding, which is omitted in Eq.~\ref{encode}.}
\begin{equation*}
    [\widetilde{E}, \widetilde{S}, \widetilde{U}_i, \widetilde{G}] = \texttt{Transformer}([\overline{E};S;U_i;G])
\label{encode}
\end{equation*}
We treat the path expansion as a multi-label classification problem with a binary classifier over $\widetilde{E}_i$, i.e., predicts whether the $i$-th entity is the next argument role for the current record and expand the path accordingly as shown in Figure~\ref{fig:tracker}.

During training, we minimize the following loss:
\begin{equation}
    \mathcal{L}_{\mathrm{record}}=-\sum_{n \in N_\mathcal{D}} \sum_{t=1}^{\left | \mathcal{E} \right |} \log P(y^n_t|n)
\label{loss:record}
\end{equation}
where $N_\mathcal{D}$ denotes the nodes set in the event records tree, and $y^n_t$ is the golden label.
If the $t$-th entity is validate for the next argument in node $n$, then $y^n_t=1$, otherwise $y^n_t=0$.

\subsection{Training}
We sum the losses coming from three sub-tasks with different weight respectively in Eq. (\ref{loss:ner}), (\ref{loss:detect}) and (\ref{loss:record}) as follows:
\begin{equation*}
    \mathcal{L}_{\mathrm{all}} = \lambda_1 \mathcal{L}_{\mathrm{ner}} + \lambda_2 \mathcal{L}_{\mathrm{detect}} + \lambda_3 \mathcal{L}_{\mathrm{record}}
\end{equation*}
More training details are shown in Appendix~\ref{appendix-training}.

%% file: main/experiments.tex
\section{Experiments}

\subsection{Dataset}
We evaluate our model on a public dataset proposed by \citet{zheng-etal-2019-doc2edag}\footnote{\url{https://github.com/dolphin-zs/Doc2EDAG/blob/master/Data.zip}}, which is constructed from Chinese financial documents.
It consists of up to $32,040$ documents which is the largest document-level EE dataset by far.
It focuses on five event types: \textit{Equity Freeze} (EF), \textit{Equity Repurchase} (ER), \textit{Equity Underweight} (EU), \textit{Equity Overweight} (EO) and \textit{Equity Pledge} (EP), with $35$ different kinds of argument roles in total.
We follow the standard split of the dataset, $25,632/3,204/3,204$ documents for training/dev/test set. 
The dataset is quite challenging, as a document has $20$ sentences and consists of $912$ tokens on average.
Besides, there are roughly $6$ sentences involved for an event record, and $29\%$ documents express multiple events.

\subsection{Experiments Setting}
In our implementation of \modelname, we use $8$ and $4$ layers Transformer \citep{NIPS2017_7181} in encoding and decoding module respectively. 
The dimensions in hidden layers and feed-forward layers are the same as previous work \citep{zheng-etal-2019-doc2edag}, i.e.,  $768$ and $1,024$.
We also use $L=3$ layers of GCN, and set dropout rate to $0.1$, batch size to $64$.
\modelname is trained using Adam \citep{DBLP:journals/corr/KingmaB14} as optimizer with $1e-4$ learning rate for $100$ epochs.
We set $\lambda_1=0.05$, $\lambda_2=\lambda_3=1$ for the loss function.

\subsection{Baselines and Metrics}
\citet{yang-etal-2018-dcfee} proposes DCFEE that extracts arguments from the identified central sentence and queries surrounding sentences for missing arguments.
The model has two variants, \textbf{DCFEE-S} and \textbf{DCFEE-M}. DCFEE-S produces one record at a time, while DCFEE-M produces multiple possible argument combinations by the closest distance from the central sentence.
Besides, \textbf{Doc2EDAG} \citep{zheng-etal-2019-doc2edag} uses transformer encoder to obtain sentence and entity embeddings, followed by another transformer to fuse cross-sentence context.
Then multiple events are extracted simultaneously.
\textbf{Greedy-Dec} is a variant of Doc2EDAG, which produces only one record greedily.

Three sub-tasks of the document-level EE are all evaluated by F1 score.
Due to limited space, we leave the results of \emph{entity extraction} and \emph{event types detection} in Appendix~\ref{appendix-evaluation}, which shows \modelname only slightly outperform Doc2EDAG, because we mainly focus on event record extraction and the methods are similar to Doc2EDAG for these two sub-tasks.
In the following, we mainly report and analyze the results of \emph{event record extraction}.

\begin{table}[t]
\centering
\scalebox{0.82}{
\begin{tabular}{lcccccc}
\toprule
\bf Model & \bf EF & \bf ER & \bf EU & \bf EO & \bf EP & \bf Overall \\
\midrule
DCFEE-S  & 46.7 & 80.0 & 47.5 & 46.7 & 56.1 & 60.3 \\
DCFEE-M  & 42.7 & 73.3 & 45.8 & 44.6 & 53.8 & 56.6 \\
Greedy-Dec  & 57.7 & 79.4 & 51.2 & 50.0 & 54.2 & 61.0 \\
Doc2EDAG  & 71.0 & 88.4 & 69.8 & 73.5 & 74.8 & 77.5 \\
\midrule
\modelname (ours) & \textbf{73.4} & \textbf{90.8} & \textbf{74.3} & \textbf{76.3} & \textbf{77.7} & \textbf{80.3} \\
\bottomrule
\end{tabular}
}
\caption{F1 scores on test set. 
\modelname achieves the best performance. We also list the results reported in \citet{zheng-etal-2019-doc2edag} in Appendix~\ref{appendix-evaluation}, and \modelname consistently outperforms other baselines. EF/ER/EU/EO/EP refer to specific event types, and Overall denotes micro F1.
}

\label{table:overall-performance}
\end{table}

\subsection{Main Results}

\textbf{Overall performance}.
The results of the overall performance on the document-level EE dataset is illustrated in Table~\ref{table:overall-performance}.
As Table~\ref{table:overall-performance} shows, our \modelname consistently outperforms other baselines, thanks to better modelling of global interactions and interdependency.
Specifically, \modelname improves $2.8$ micro F1 compared with the previous state-of-the-art, Doc2EDAG, especially $4.5$ improvement in \textit{Equity Underweight} (EU) event type.

\begin{table}[t]
\centering
\scalebox{1.0}{
\begin{tabular}{lcccc}
\toprule
\bf Model & \bf \uppercase\expandafter{\romannumeral1} & \bf \uppercase\expandafter{\romannumeral2} & \bf \uppercase\expandafter{\romannumeral3} & \bf \uppercase\expandafter{\romannumeral4} \\
\midrule
DCFEE-S & 64.6 & 70.0 & 57.7 & 52.3 \\
DCFEE-M & 54.8 & 54.1 & 51.5 & 47.1 \\
Greedy-Dec & 67.4 & 68.0 & 60.8 & 50.2\\
Doc2EDAG & 79.6 & 82.4 & 78.4 & 72.0\\
\midrule
\modelname (ours) & \textbf{81.9} & \textbf{85.7} & \textbf{80.0} & \textbf{75.7} \\
\bottomrule
\end{tabular}
}
\caption{F1 scores on four sets with growing average number of involved sentences for records (increases from \uppercase\expandafter{\romannumeral1} to \uppercase\expandafter{\romannumeral4}). The highest improvement of \modelname comes from event records involving the most sentences (Set  \uppercase\expandafter{\romannumeral4}) by $3.7$ F1 score compared with Doc2EDAG.
}
\label{table:cross-sentence}
\end{table}

\textbf{Cross-sentence records scenario.}
There are more than $99.5\%$ records of the test set are cross-sentence event records, and the extraction becomes gradually more difficult as the number of their involved sentences grows.
To verifies the effectiveness of \modelname to capture cross-sentence information, we first calculate the average number of sentences that the records involve for each document, and sort them in ascending order.
Then we divide them into four sets \uppercase\expandafter{\romannumeral1}/\uppercase\expandafter{\romannumeral2}/\uppercase\expandafter{\romannumeral3}/\uppercase\expandafter{\romannumeral4} with equal size.
Documents in Set. \uppercase\expandafter{\romannumeral4} is considered to be the most challenging as it requires the most number of sentences to successfully extract records.
As Table~\ref{table:cross-sentence} shows, \modelname consistently outperforms Doc2EDAG, especially on the most challenging Set. \uppercase\expandafter{\romannumeral4} that involves the most sentences, by $3.7$ F1 score.
It suggests that \modelname can well capture global context and mitigate the arguments-scattering challenge, with the help of the heterogeneous graph interaction network.

\begin{table*}[htbp]
\centering
\begin{tabular}{lcccccccccccc}
\toprule
\multirow{2}{*}{\bf Model} & \multicolumn{2}{c}{\bf EF} & \multicolumn{2}{c}{\bf ER} & \multicolumn{2}{c}{\bf EU} & \multicolumn{2}{c}{\bf EO} & \multicolumn{2}{c}{\bf EP} & \multicolumn{2}{c}{\bf Overall}\\ 
\cmidrule(lr){2-3} 
\cmidrule(lr){4-5}
\cmidrule(lr){6-7}
\cmidrule(lr){8-9}
\cmidrule(lr){10-11}
\cmidrule(lr){12-13}
~ & \bf S. & \bf M. & \bf S. &  \bf M. & \bf S. &  \bf M. & \bf S. &  \bf M. & \bf S. &  \bf M. & \bf S. &  \bf M. \\
\midrule
DCFEE-S & 55.7 & 38.1 & 83.0 & 55.5 & 52.3 & 41.4 & 49.2 & 43.6 & 62.4 & 52.2 & 69.0 & 50.3\\
DCFEE-M & 45.3 & 40.5 & 76.1 & 50.6 & 48.3 & 43.1 & 45.7 & 43.3 & 58.1 & 51.2 & 63.2 & 49.4 \\
Greedy-Dec & 74.0 & 40.7 & 82.2 & 50.0 & 61.5 & 35.6 & 63.4 & 29.4 & 78.6 & 36.5 & 77.8 & 37.0 \\
Doc2EDAG & 79.7 & 63.3 & 90.4 & 70.7 & 74.7 & 63.3 & 76.1 & 70.2 & 84.3 & 69.3 & 81.0 & 67.4\\
\midrule
\modelname (ours) & \textbf{81.9} & \textbf{65.9} & \textbf{93.0} & \textbf{71.7} & \textbf{82.0} & \textbf{64.1} & \textbf{80.9} & \textbf{70.6} & \textbf{85.0} & \textbf{73.5} & \textbf{87.6} & \textbf{72.3} \\
\bottomrule
\end{tabular}
\caption{F1 scores on single-record (S.) and multi-record (M.) sets.}
\label{table:single-vs-multi}
\end{table*}

\textbf{Multiple records scenario.}
\modelname introduces the \textit{tracker} to make use of global interdependency among event records, which is important in multiple records scenario.
To illustrate its effectiveness, we divide the test set into single-record set (S.) containing documents with one record, and multi-record set (M.) containing those with multiple records.
As shown in Table.~\ref{table:single-vs-multi}, F1 score on M. is much lower than that on S., indicating it is challenging to extract multiple records.
However, \modelname still surpasses other strong baselines by $ 4.9 \sim 35.3$ on multi-record set (M.).
This is because \modelname is aware of other records through the $Tracker$ module, and leverage the interdependency information to improve the performance\footnote{
\citet{nguyen-etal-2016-joint} maintain three binary matrices to memorize entities and events states. Although they aim at sentence-level EE that contains fewer entities and event records, it would be also interesting to compare with them and we leave it as future work.}.

\begin{table}[t]
\centering
\scalebox{1.0}{
\begin{tabular}{lc|cccc}
\toprule
\bf Model & \bf F1 & \bf \uppercase\expandafter{\romannumeral1} & \bf \uppercase\expandafter{\romannumeral2} & \bf \uppercase\expandafter{\romannumeral3} & \bf \uppercase\expandafter{\romannumeral4} \\
\midrule
\modelname & \textbf{80.3} & \textbf{81.9} & \textbf{85.7} & \textbf{80.0} & \textbf{75.7} \\
- S-S & -1.4 & -0.9 & -0.1 & -1.9 & -2.3 \\
- S-M & -1.0 & -1.6 & -1.7 & -0.7 & -0.7 \\
- M-M$_{\mathrm{intra}}$ & -1.3 & -0.5 & -1.4 & -2.4 & -1.5 \\
- M-M$_{\mathrm{inter}}$ & -1.1 & -0.5 & -1.6 & -1.4 & -1.7 \\
- Graph & -2.0 & -1.8 & -1.5 & -2.0 & -2.5\\
\bottomrule
\end{tabular}
}
\caption{The decrease of F1 scores on ablation study for \modelname's heterogeneous graph interaction network. Removing the heterogeneous graph leads to significant drop on F1, especially for records involving the most sentences (i.e., $-2.5$ F1 on Set \uppercase\expandafter{\romannumeral4}).
}
\label{table:graph}
\end{table}

\begin{table}[t]
\centering
\scalebox{1.0}{
\begin{tabular}{cccc|cc}
\toprule
\bf Model & \bf P & \bf R & \bf F1 & \bf S. & \bf M. \\
\midrule
\modelname & \textbf{82.3} & \textbf{78.4} & \textbf{80.3} & \textbf{87.6} & \textbf{72.3} \\
\modelname-OT & -0.6 & -0.4 & -0.5 & -0.8 & -0.7 \\
\modelname-OP & -1.0 & -1.6 & -1.2 & -1.0 & -1.5 \\
\modelname-NT & -2.8 & +0.1 & -1.3 & -1.3 & -1.5 \\
\bottomrule
\end{tabular}
}
\caption{Performance of \modelname on ablation study for the $Tracker$ module. The removal of the \textit{Tracker} (\modelname-NT) brings about higher F1 decrease on M. than that on S..
S.: Single-record set, M.: Multi-record set.
}
\label{table:tracker}
\end{table}

\begin{figure}[t]
    \centering
    \includegraphics[width=1.0\linewidth]{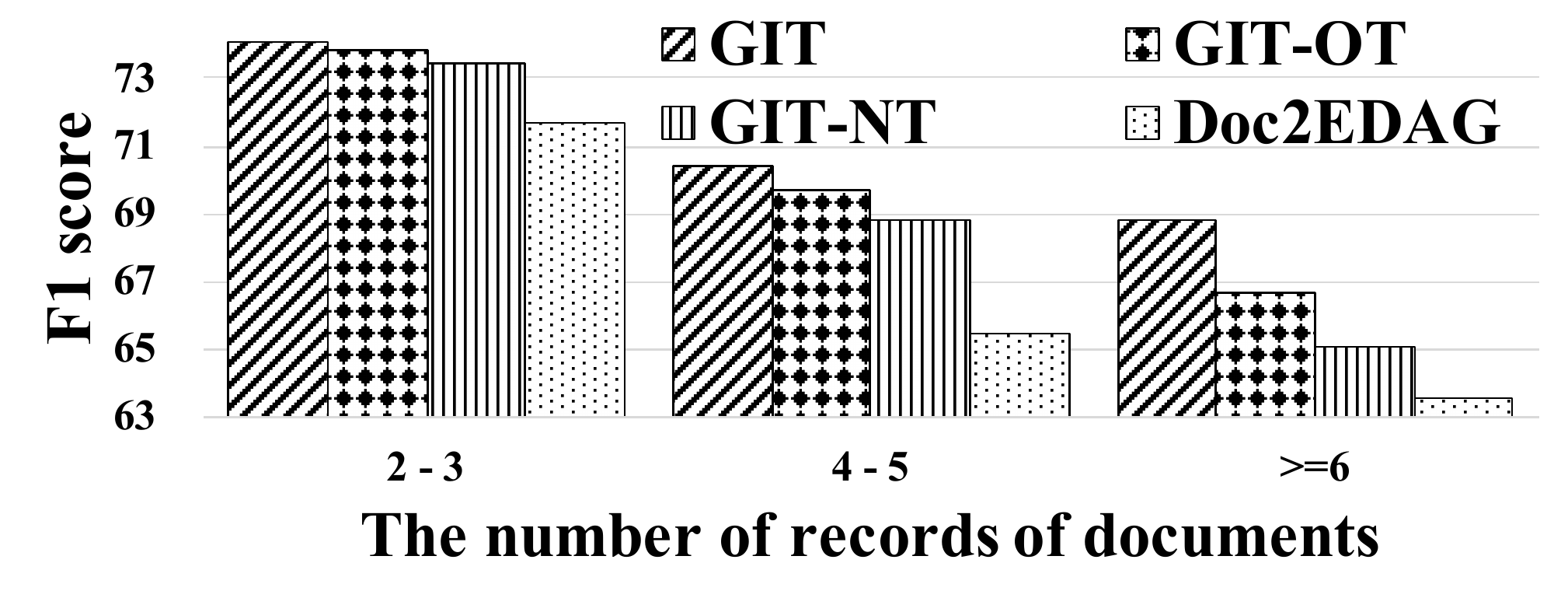}
    \caption{F1 scores on documents with different number of event records. The F1 gap between w/ (\modelname) and w/o \textit{Tracker} (\modelname-NT) becomes wider as the number of event records of documents increases. }
    \label{fig:multi-detail}
\end{figure}

\begin{figure*}[t]
    \centering
    \includegraphics[width=1.0\linewidth]{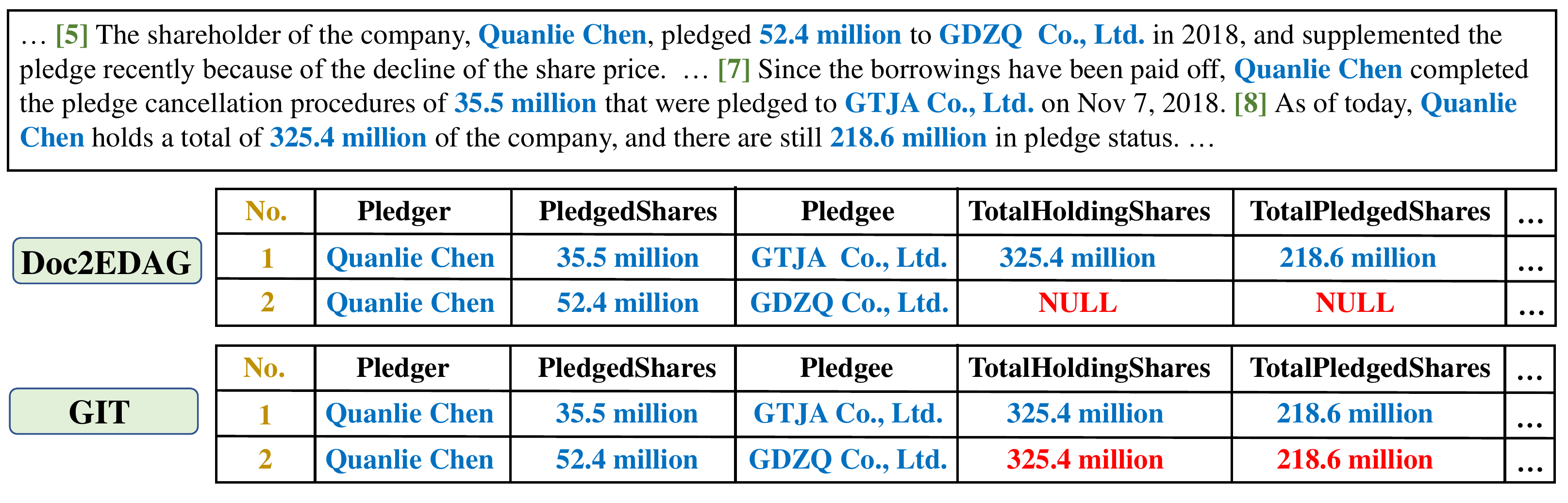}
    \caption{The case study of our proposed \modelname and Doc2EDAG, with their key prediction difference colored in red.
    Related entities are colored in blue.
    \modelname successfully extract \textit{TotalHoldingShares} and \textit{TotalPledgedShares} for Record $2$, while Doc2EDAG fails.
    The complete content are provided in Appendix~\ref{appendix-case-study}. }
    \label{fig:case-study}
\end{figure*}

\subsection{Analysis}
\label{sec:analysis}
We conduct further experiments to analyze the  key modules in \modelname more deeply. 

\textbf{On the effect of heterogeneous graph interaction network}. 
The heterogeneous graph we constructed contains four types of edges.
To explore their functions, we remove one type of edges at a time, and remove the whole graph network finally.
Results are shown in Table~\ref{table:graph}, including micro F1 and F1 on the four sets, which are divided by the number of involved sentences for records as we did before.
The micro F1 would decreases $1.0\sim1.4$ without a certainty type of edge.
Besides, removing the whole graph causes an significant drop by $2.0$ F1, especially for Set \uppercase\expandafter{\romannumeral4} by $2.5$, which requires the most number of sentences to extract the event record.
It demonstrates that the graph interaction network helps improve the performance, especially on records involving many sentences, and all kinds of edges play an important role for extraction.

\textbf{On the effect of \textit{Tracker} module}.
\modelname can leverage interdependency among records based on the information of other event records tracked by \textit{Tracker}.
To explore its effect, firstly, we remove the global interdependency information between records of different event types, by clearing the global memory whenever we extract events for another new event type (\modelname-\textbf{O}wn \textbf{T}ype).
Next, we remove all the tracking information except the own path for a record, to explore whether the tracking of other records makes effect indeed (\modelname-\textbf{O}wn \textbf{P}ath).
Finally, we remove the whole \textit{Tracker} module (\modelname-\textbf{N}o \textbf{T}racker).
As Table~\ref{table:tracker} shows, the F1 in \modelname-OT/\modelname-OP decreases by $0.5$/$1.2$, suggesting the interdependency among records of both the same and different event types do play an essential role.
Besides, their F1 decrease in M. by $0.7$/$1.5$ are more than those in S. by $0.8$/$1.0$, verifying the effectiveness of the \textit{Tracker} in multi-event scenarios.
Moreover, the performances are similar between \modelname-OP and \modelname-NT, which also provides evidence that other records do help.
We also reveal F1 on documents with different number of records in Figure~\ref{fig:multi-detail}.
The gap between models with or without \textit{Tracker} raises as the number of records increases, which validates the effectiveness of our \textit{Tracker}.

\subsection{Case Study}
Figure~\ref{fig:case-study} demonstrates a case of the predictions of Doc2EDAG and \modelname for \textit{Equity Pledge} (EP) event types.
The \textit{TotalHoldingShares} and \textit{TotalPledgedShares} information lies in Sentence $8$, while the \textit{PledgedShares} and \textit{Pledgee} information for Record $2$ lies in Sentence $5$.
Though Doc2EDAG fails to extract these arguments in Record $2$ (colored in red), \modelname succeeds because it can capture interactions between long-distance sentences, and utilize the information of Record $1$ (\textit{325.4 million} and \textit{218.6 million}) thanks to the \textit{Tracker} model.

%% file: main/related.tex
\section{Related Work}

\textbf{Sentence-level Event Extraction}. 
Previous approaches mainly focus on sentence-level event extraction.
\citet{chen-etal-2015-event} propose a neural pipeline model that identifies triggers first and then extracts argument roles.
\citet{nguyen-etal-2016-joint} use a joint model to extract triggers and argument roles simultaneously.
Some studies also utilize dependency tree information~\citep{liu-etal-2018-jointly, yan-etal-2019-event}.
To utilize more knowledge, some studies leverage 
document context \citep{chen-etal-2018-collective, zhao-etal-2018-document}, pre-trained language model \citep{yang-etal-2019-exploring-pre}, and explicit external knowledge \citep{DBLP:conf/aaai/Liu0019, tong-etal-2020-improving} such as WordNet \citep{DBLP:journals/cacm/Miller95}.
\citet{du-cardie-2020-event} also try to extract events in a Question-Answer way.
These studies usually conduct experiments on sentence-level event extraction dataset,  ACE05~\citep{ace05}.
However, it is hard for the sentence-level models to extract multiple qualified events spanning across sentences, which is more common in real-world scenarios.

\textbf{Document-level Event Extraction}.
Document-level EE has attracted more and more attention recently.
\citet{yang-mitchell-2016-joint} use well-defined features to handle the event-argument relations across sentences, which is, unfortunately, quite nontrivial.
\citet{yang-etal-2018-dcfee} extract events from a central sentence and find other arguments from neighboring sentences separately.
Although \citet{zheng-etal-2019-doc2edag} use Transformer to fuse sentences and entities, interdependency among events is neglected.
\citet{du-cardie-2020-document} try to encode the sentences in a multi-granularity way and \citet{DBLP:journals/corr/abs-2008-09249} leverage a seq2seq model.
They conduct experiments on MUC-4~\citep{muc} dataset with $1,700$ documents and $5$ kinds of entity-based arguments, and it is formulated as a table-filling task, coping with single event record of single event type.
However, our work is different from these studies in that
a) we utilize heterogeneous graph to model the global interactions among sentences and mentions to capture cross-sentence context,
b) and we leverage the global interdependency through \textit{Tracker} to extract multiple event records of multiple event types.

%% file: main/conclusion.tex
\section{Conclusion}

Although promising in practical application, document-level EE still faces some challenges such as arguments-scattering phenomenon and multiple correlated events expressed by a single document.
To tackle the challenges, we introduce Heterogeneous \textbf{G}raph-based \textbf{I}nteraction Model with a \textbf{T}racker (\modelname).
\modelname uses a heterogeneous graph interaction network to model global interactions among sentences and entity mentions.
\modelname also uses a \textit{Tracker} to track the extracted records to consider global interdependency during extraction.
Experiments on large-scale public dataset \citep{zheng-etal-2019-doc2edag} show \modelname outperforms previous state-of-the-art by $2.8$ F1.
Further analysis verifies the effectiveness of \modelname especially in cross-sentence events extraction and multi-event scenarios.

%% file: main/appendix.tex
\appendix

\section{Training Details}
\label{appendix-training}
To mitigate the error propagation due to the gap between training and inference phrase (i.e., the extracted entities are ground truth during training but predicted results during inference), we adopt scheduled sampling strategy \citep{10.5555/2969239.2969370} as \citet{zheng-etal-2019-doc2edag} did.
We gradually switch the entity extraction results from golden label to what the model predicts on its own. 
Specifically, from epoch $10$ to epoch $20$, we linearly increase the proportion of predicted entity results from $0\%$ to $100\%$.
We implement \modelname under PyTorch \citep{paszke2017automatic} and DGL \citep{wang2019dgl} based on codes provided by \citet{zheng-etal-2019-doc2edag}.

All the experiments (including the baselines) are run with the same $8$ Tesla-V100 GPUs and the same version of python dependencies to ensure the fairness.

Hyperparameters trials are listed in Table~\ref{tab:hyperparam}. 
The value of hyperparameters we finally adopted are in bold. 
Note that we do not tune all the hyperparameters, and make little effort to select the best hyperparameters for our \modelname.

\begin{table}[htbp]
\small
\centering
\begin{tabular}{lc}
\hline
\bf Hyperparameters & \bf Value \\
\hline
Batch Size &  32, \textbf{64} \\
Learning Rate  & \textbf{0.0001}  \\
Dropout &  \textbf{0.1}\\
Layers of GCN & 1, 2, \textbf{3}, 4, 5 \\
Number of Epochs & \textbf{100} \\
$\lambda_1$ & \textbf{0.05} \\
$\lambda_2$ & \textbf{1.00} \\
$\lambda_3$ & \textbf{1.00} \\
Gradient Accumulation Steps & \textbf{8} \\
Layers of Transformer in Entity Extractor & \textbf{8} \\
Layers of Transformer in Decoder Module & \textbf{4} \\
\hline \hline
Hyperparameter Search Trials & 10 \\
\hline
\end{tabular}
\caption{Hyperparameters for our proposed \modelname.
}
\label{tab:hyperparam}
\end{table}

We choose the final checkpoints for test according to the Micro F1 performance on the dev set. 
Table~\ref{table:dev} illustrates the best epoch in which the model achieves the highest Micro F1 on the dev set and their according F1 score.

\begin{table}[htbp]
\centering
\begin{tabular}{lcccccc}
\hline
\bf Model & \bf P & \bf R & \bf F1 \\
\hline
DCFEE-S  & 86.5 & 88.6 & 87.6 \\
DCFEE-M  & 86.6 & 89.0 & 87.8 \\
Greedy-Dec & 87.5 & 89.8 & 88.6 \\
Doc2EDAG  &  88.0 & 90.0 & 89.0\\
\hline
\modelname (ours) & 85.8 & 92.6 & 89.1 \\
\hline
\end{tabular}
\caption{Results of \textbf{entity extraction} sub-task on the test set. The performance of different models are similar, for the reason that they all utilize the same structure and methods to extract entities.
}
\label{table:ner}
\end{table}

\begin{table*}[htbp]
\centering
\begin{tabular}{lcccccc}
\hline
\bf Model & \bf EF & \bf ER & \bf EU & \bf EO & \bf EP & \bf Overall \\
\hline
DCFEE-S  & 81.5 & 94.0 & 82.3 & 85.7 & 93.8 & 91.4 \\
DCFEE-M  & 79.8 & 92.4 & 78.9 & 84.2 & 92.9 & 90.0 \\
Greedy-Dec  & 99.3 & 99.9 & 96.8 & 95.4 & 99.6 & 99.0 \\
Doc2EDAG  & 99.0 & 99.8 & 96.8 & 94.1 & 99.5 & 98.9 \\
\hline
\modelname (ours) & 98.8 & 99.8 & 97.9 & 96.6 & 99.6 & 99.2 \\
\hline
\end{tabular}
\caption{F1 scores results of \textbf{event types detection} sub-task on the test set. All the models obtains more than 90.0 micro F1 score. \modelname slightly outperform Doc2EDAG.
}
\label{table:triggering}
\end{table*}

\begin{table*}[htbp]
\centering
\begin{tabular}{lccccccc}
\hline
\bf Model & \bf Best Epoch & \bf EF & \bf ER & \bf EU & \bf EO & \bf EP & \bf Overall \\
\hline
DCFEE-S  & 86 & 51.3 & 73.0 & 44.1 & 51.4 & 58.6 & 58.7 \\
DCFEE-M  & 87 & 52.5 & 69.1 & 43.9 & 47.2 & 55.9 & 55.8 \\
Greedy-Dec  & 90 & 57.5 & 76.0 & 55.1 & 49.3 & 57.0 & 59.1 \\
Doc2EDAG  & 89 & 75.2 & 85.2 & 71.6 & 80.0 & 77.9 & 78.7 \\
\hline
\modelname (ours) & 89 & \bf 78.3 & \bf 87.6 & \bf 74.7 & \bf 80.9 & \bf 79.8 & \bf 80.7 \\
\hline
\end{tabular}
\caption{The best epoch in which the models achieve the highest \textbf{micro F1 score on the dev set} and the corresponding performance.
}
\label{table:dev}
\end{table*}

\begin{figure*}[h]
    \centering
    \includegraphics[width=1.0\linewidth]{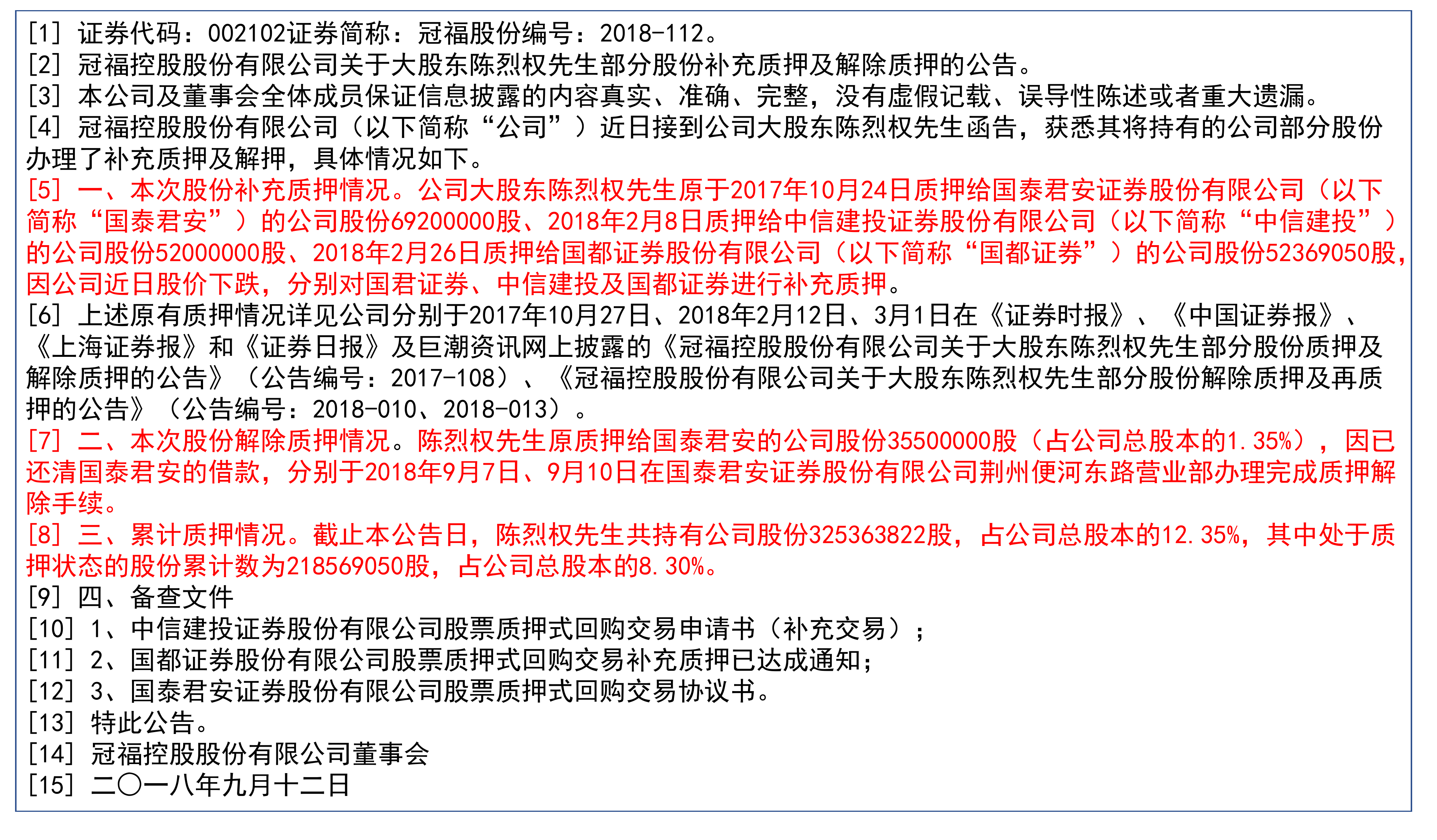}
    \caption{The original complete document corresponding to the case study in Figure~\ref{fig:case-study}. Sentences in red color are presented in Figure~\ref{fig:case-study}.}
    \label{fig:case-study-origin}
\end{figure*}

\section{Additional Evaluation Results}
\label{appendix-evaluation}
We have showed the evaluation results of event records extraction in the paper for document-level event extraction.
In this section, we also illustate the results of entity extraction in Table.~\ref{table:ner} and event types detection in Table.~\ref{table:triggering}.
Moreover, the comprehensive results of event record extraction is shown in Table.~\ref{table:all}, including results reported in \citet{zheng-etal-2019-doc2edag} with precison, recall and F1 score.

\begin{table*}[htbp]
\centering
\begin{adjustbox}{angle=90}
\begin{tabular}{lcccccccccccccccccc}
\hline
\multirow{2}{*}{\bf Model} & \multicolumn{3}{c}{\bf EF} & \multicolumn{3}{c}{\bf ER} & \multicolumn{3}{c}{\bf EU} & \multicolumn{3}{c}{\bf EO} & \multicolumn{3}{c}{\bf EP} & \multicolumn{3}{c}{\bf Overall}\\ 
\cmidrule(lr){2-4} 
\cmidrule(lr){5-7}
\cmidrule(lr){8-10}
\cmidrule(lr){11-13}
\cmidrule(lr){14-16}
\cmidrule(lr){17-19}
~ & \bf P & \bf R & \bf F1  & \bf P & \bf R & \bf F1 & \bf P & \bf R & \bf F1 & \bf P & \bf R & \bf F1 & \bf P & \bf R & \bf F1 & \bf P & \bf R & \bf F1 \\
\hline
DCFEE-S$^\diamondsuit$  & 66.0 & 41.6 & 51.1 & 84.5 & 81.8 & 83.1 & 62.7 & 35.4 & 45.3 & 51.4 & 42.6 & 46.6 & 64.3 & 63.6 & 63.9 & - & - & - \\
DCFEE-M$^\diamondsuit$  & 51.8 & 40.7 & 45.6 & 83.7 & 78.0 & 80.8 & 49.5 & 39.9 & 44.2 & 42.5 & 47.5 & 44.9 & 59.8 & 66.4 & 62.9 & - & - & - \\
Greedy-Dec$^\diamondsuit$  & \bf 79.5 & 46.8 & 58.9 & 83.3 & 74.9 & 78.9 & 68.7 & 40.8 & 51.2 & 69.7 & 40.6 & 51.3 & \bf 85.7 & 48.7 & 62.1 & - & - & - \\
Doc2EDAG$^\diamondsuit$  & 77.1 & 64.5 & 70.2 & 91.3 & 83.6 & 87.3 & 80.2 & 65.0 & 71.8 & \bf 82.1 & 69.0 & 75.0 & 80.0 & 74.8 & 77.3 & - & - & - \\
\hline
DCFEE-S$^\spadesuit$  & 61.1 & 37.8 & 46.7 & 84.5 & 76.0 & 80.0 & 60.8 & 39.0 & 47.5 & 46.9 & 46.5 & 46.7 & 64.2 & 49.8 & 56.1 & 67.7 & 54.4 & 60.3 \\
DCFEE-M$^\spadesuit$  & 44.6 & 40.9 & 42.7 & 75.2 & 71.5 & 73.3 & 51.4 & 41.4 & 45.8 & 42.8 & 46.7 & 44.6 & 55.3 & 52.4 & 53.8 & 58.1 & 55.2 & 56.6 \\
Greedy-Dec$^\spadesuit$  & 78.5 & 45.6 & 57.7 & 83.9 & 75.3 & 79.4 & 69.0 & 40.7 & 51.2 & 64.8 & 40.6 & 50.0 & 82.1 & 40.4 & 54.2 & 80.4 & 49.1 & 61.0 \\
Doc2EDAG$^\spadesuit$  & 78.7 & 64.7 & 71.0 & 90.0 & 86.8 & 88.4 & 80.4 & 61.6 & 69.8 & 77.2 & 70.1 & 73.5 & 76.7 & 73.0 & 74.8 & 80.3 & 75.0 & 77.5 \\
\hline
\modelname (ours)$^\spadesuit$ & 78.9 & \bf 68.5 & \bf 73.4 & \bf 92.3 & \bf 89.2 & \bf 90.8 & \bf 83.9 & \bf 66.6 & \bf 74.3 & 80.7 & \bf 72.3 & \bf 76.3 & 78.6 & \bf 76.9 & \bf 77.7 & \bf 82.3 & \bf 78.4 & \bf 80.3 \\
\hline
\end{tabular}
\end{adjustbox}
\caption{Comprehensive results of event record extraction. Results with $\diamondsuit$ are results reported in \citet{zheng-etal-2019-doc2edag}. Results with are $\spadesuit$ results we implement on our own. Our \modelname consistently outperform other baselines.
}
\label{table:all}
\end{table*}

\section{Complete Document for the Examples}
\label{appendix-running-example}
\label{appendix-case-study}

\begin{figure*}[ht]
    \centering
    \includegraphics[width=0.9\linewidth]{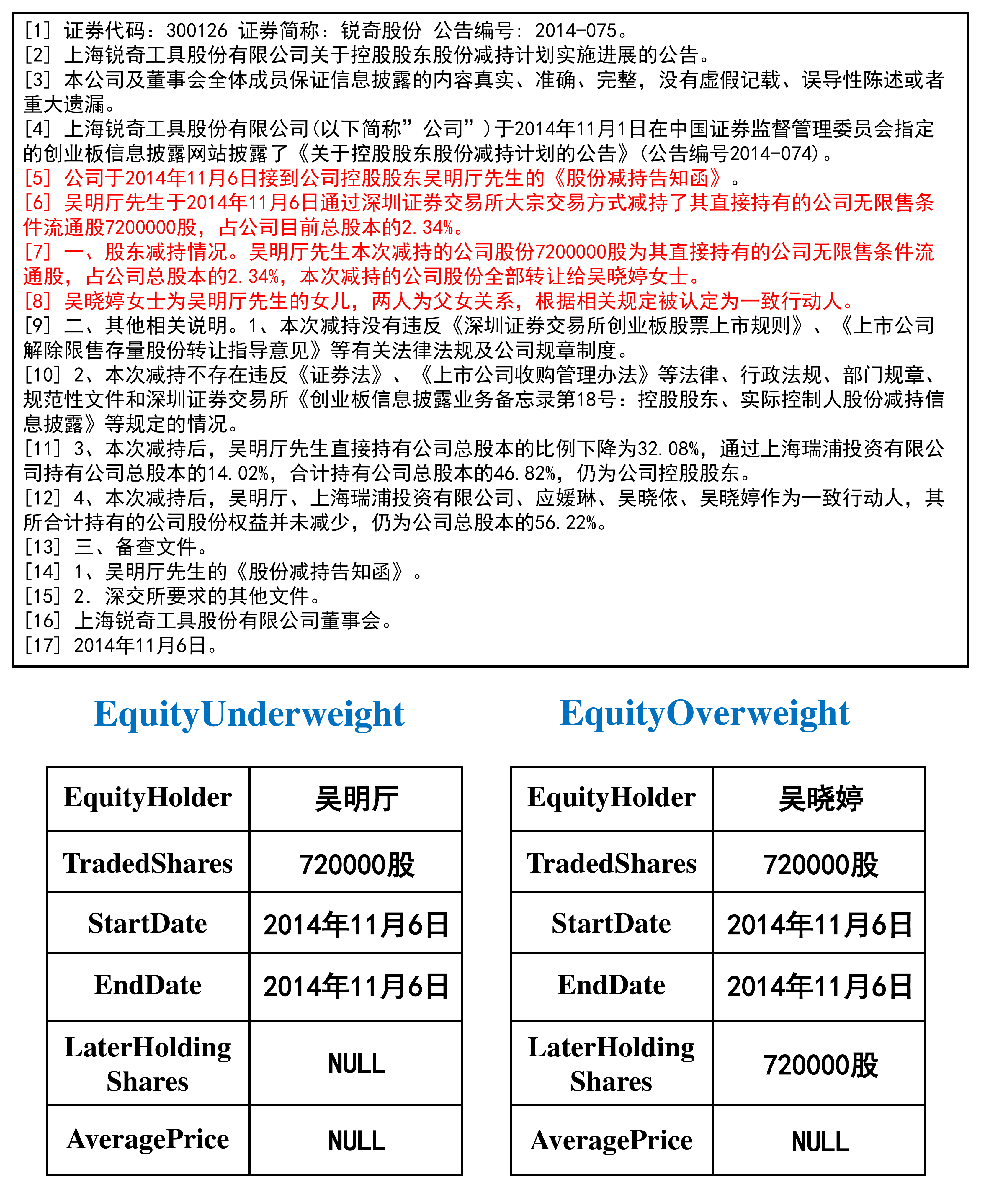}
    \caption{The original complete document corresponding to the running example in Figure~\ref{fig:running-example}. Sentences in red color are presented in Figure~\ref{fig:running-example}.}
    \label{fig:running-example-origin}
\end{figure*}

We show an example document in Figure~\ref{fig:running-example} in the paper.
To better illustrate, we translate it from Chinese into English and make some simplication.
Here we present the original complete document example in Figure~\ref{fig:running-example-origin}. 
For the specific meanings of argument roles, we recommend readers to refer to \citep{zheng-etal-2019-doc2edag}.

We also demonstrate an case study in Figure~\ref{fig:case-study} in the paper.
Now we also show its original Chinese version in  Figure~\ref{fig:case-study-origin}.

